\documentclass[conference,letterpaper]{IEEEtran}
\IEEEoverridecommandlockouts
\usepackage{cite}
\usepackage{amsmath,amssymb,amsfonts}
\usepackage{algorithmic}
\usepackage{graphicx}
\usepackage{textcomp}
\usepackage[OT1]{fontenc}
\usepackage{pifont}
\usepackage{multirow}
\usepackage{threeparttable}
\usepackage{booktabs}

\usepackage[dvipsnames,svgnames]{xcolor, colortbl}

\definecolor{LightCyan}{rgb}{0.88,1,1}
\def\BibTeX{{\rm B\kern-.05em{\sc i\kern-.025em b}\kern-.08em
    T\kern-.1667em\lower.7ex\hbox{E}\kern-.125emX}}
\begin{document}

\title{Towards a Multi-Agent Vision-Language System for Zero-Shot Novel Hazardous Object Detection for Autonomous Driving Safety\\
}

\author{
\IEEEauthorblockN{
Shashank Shriram$^1$, Srinivasa Perisetla$^1$, Aryan Keskar$^1$, Harsha Krishnaswamy$^1$,\\ Tonko Emil Westerhof Bossen$^{1,2}$, Andreas Møgelmose$^2$, Ross Greer$^1$}
\IEEEauthorblockA{
\textit{Machine Intelligence, Interaction, and Imagination (Mi$^3$) Laboratory}\\
\textit{$^1$University of California, Merced} \\
\textit{$^2$Aalborg Universitet}
}
}

\maketitle

\begin{abstract}
Detecting anomalous hazards in visual data, particularly in video streams, is a critical challenge in autonomous driving. Existing models often struggle with unpredictable, out-of-label hazards due to their reliance on predefined object categories. In this paper, we propose a multimodal approach that integrates vision-language reasoning with zero-shot object detection to improve hazard identification and explanation. Our pipeline consists of a Vision-Language Model (VLM), a Large Language Model (LLM), in order to detect hazardous objects within a traffic scene. We refine object detection by incorporating OpenAI’s CLIP model to match predicted hazards with bounding box annotations, improving localization accuracy. To assess model performance, we create a ground truth dataset by denoising and extending the foundational COOOL (Challenge-of-Out-of-Label) anomaly detection benchmark dataset with complete natural language descriptions for hazard annotations. We define a means of hazard detection and labeling evaluation on the extended dataset using cosine similarity. This evaluation considers the semantic similarity between the predicted hazard description and the annotated ground truth for each video. Additionally, we release a set of tools for structuring and managing large-scale hazard detection datasets. Our findings highlight the strengths and limitations of current vision-language-based approaches, offering insights into future improvements in autonomous hazard detection systems. Our models, scripts, and data can be found at https://github.com/mi3labucm/COOOLER.git
\end{abstract}


\section{Introduction}

The rise of autonomous vehicles underscores the challenge of ensuring safety in diverse and unpredictable environments. While significant progress has been made in handling standard road conditions, autonomous systems often struggle with rare or unconventional hazards that fall outside typical training distributions \cite{wang2019parametric, liu2020understanding, liang2024aide}. Addressing these limitations is crucial for enhancing the robustness and real-world deployment of autonomous driving technology.

This research advances out-of-label hazard detection by integrating Vision-Language Models (VLMs) such as ViLA \cite{lin2023vila, lin2024vila, liu2024nvila} and OmniVLM \cite{chen2024omnivlm}, alongside Large Language Models (LLMs) like OpenAI's ChatGPT-4 family of models, to identify rare and unconventional traffic hazards. We then incorporate OpenAI's CLIP model in order to match predicted hazards with bounding box data to improve localization accuracy. Unlike traditional datasets that emphasize common driving conditions, our approach targets anomalies, which include debris, animals, and erratic pedestrian behaviors that pose critical but often overlooked risks.

Building from the landmark COOOL (Challenge-of-Out-of-Label) benchmark\footnote{To clarify our terminology, we use \textit{benchmark} to refer to a standardized dataset, task, and means of evaluation. While COOOL provides unlabeled data, in its current form, the benchmark does not provide annotated hazards and only allows black-box evaluation based on string subsequence matching to an unknown string set, necessitating an open-set language extension for validating our hazard recognition methods.} dataset \cite{alshami2024coool}, we introduce a dataset extension and performance measurement method to validate our approach to anomaly recognition. This dataset extension includes denoising and preprocessing of the COOOL videos, hazard annotations, and definition of an open-set method of evaluation to use in our experiments to validate our methods of anomaly detection and description. The dataset within this benchmark includes 200 short video clips capturing diverse traffic scenes featuring anomalies and hazards and annotated bounding boxes for all objects within the scene. We name our augmentation of the dataset \textit{COOOLER} (COOOL with Expanded Representations); our augmentation of the COOOL benchmark includes the following for each video scene:

\begin{enumerate}
    \item Removal of video watermarks for more naturalistic visual driving scenes
    \item Denoising and deblurring of the traffic scenes for enhanced video quality via NAFNet \cite{chen2022simple}
    \item Human annotations for the descriptions of each Hazardous object within the video sequence.
    \item Publicly-available, open-set methods of evaluation to human-annotated ground truth hazard descriptions which include cosine similarity and embedding-based caption analysis
\end{enumerate}

By bridging vision-language understanding with improved hazard detection, this research enhances the ability of autonomous vehicles to recognize real-world anomalies through the use of VLMs, LLMs, CLIP and a new evaluation benchmark through COOOLER, ultimately improving safety and reliability in unpredictable autonomous driving conditions.

\section{Related Research}

\subsection{Vision Language Models}

In recent years, vision-language models have made significant advancements in bridging the gap between visual perception and linguistic reasoning. These models, including ViLA \cite{liu2024nvila}, OmniVLM \cite{chen2024omnivlm}, GPT-4V \cite{yang2023dawn}, and CLIP \cite{radford2021learning}, have demonstrated strong capabilities in tasks such as image captioning, question answering, and zero-shot object recognition. The broader class of vision-language models has improved scene understanding by enhancing the ability to describe, interpret, and analyze visual inputs, making them increasingly relevant in domains like autonomous driving, where accurate perception and contextual reasoning are essential.
ViLA, for example, is particularly effective at aligning visual information with textual descriptions, making it well-suited for tasks requiring detailed scene interpretation. This allows it to generate fine-grained image captions, answer complex questions about visual content, and provide richer contextual understanding of driving environments. OmniVLM, on the other hand, prioritizes robust generalization, allowing it to interpret diverse driving environments effectively. GPT-4V advances the field by generating rich textual descriptions of complex visuals, which improves scene summarization and anomaly explanation. CLIP, a model based on contrastive learning, facilitates zero-shot object detection by linking textual descriptions with visual features, minimizing the need for large labeled datasets.
Beyond these models, advancements in image and video processing have further enhanced the quality of vision-language tasks. However, despite these advancements, significant challenges remain, particularly in anomaly detection where models must recognize subtle deviations in dynamic environments. Future developments in VLMs will likely focus on refining their ability to detect rare hazards, improving their robustness in novel situations, and strengthening their capacity for contextual understanding in real-world autonomous systems.

\subsection{Autonomous Driving Benchmarks}

The creation of benchmarks such as nuScenes \cite{caesar2020nuscenes}, Cityscapes \cite{cordts2016cityscapes}, and KITTI \cite{geiger2013vision} have contributed to the development of autonomous driving technologies. These datasets provide high-quality annotations for tasks like segmentation, trajectory prediction, and object detection; however, they primarily focus on well-defined, common scenarios. For example, the nuScenes dataset has become a widely-used resource in autonomous driving research due to its multimodal data collection. The dataset consists of 1,000 annotated scenes, lasting 20 seconds each, comprising a total of 1.4 million images and detailed 3D bounding box annotations. These annotations cover various conditions, including low light, rain, and dense traffic, making it highly versatile. However, the nuScenes dataset has limitations in addressing unconventional hazards or rare occurrences. nuScenes and similar datasets such as KITTI and Cityscapes primarily focus on predefined categories that fail to capture the ``novelty problem`` where autonomous systems encounter anomalous events.

This dataset aims to advance the prediction of hazards by emphasizing the importance of reasoning and logical inference. The DHPR dataset includes 15,000 annotated dash-cam images with hazard descriptions, bounding boxes, and car speeds. This dataset uses vision and language models to establish a baseline for integrating reasoning into hazard prediction systems. However, the DHPR dataset’s reliance on signal images prevents temporal reasoning which is essential for understanding dynamic driving scenarios. These limitations highlight the need for datasets and methods that integrate temporal and contextual factors for comprehensive hazard analysis.


\subsection{Existing Approaches to Hazard Detection}

Here, we describe existing approaches to hazard detection and identification utilizing visual information. Recent work in hazard detection has explored various approaches, including multimodal and vision language based frameworks.  Charoenpitaks et al. address the challenge of predicting driving hazards using multimodal AI \cite{charoenpitaks2024exploring}; they introduce visual abductive reasoning (VAR) as an approach to infer accidents from dash-cam images, and demonstrate results on the Driving Hazard Prediction and Reasoning dataset (DHPR). 

Zhouxiang and Petrosian \cite{zhouxiang2025driver} propose a multimodal driver assistance system that integrates road conditions, facial video analysis, and additional sensory inputs to improve hazard detection. Their apprroach focuses on fusing multiple data sources to enhance situational awareness. 
Chen et al. \cite{chen2025insight} introduce a hierarchical VLM framework to enhance hazard detection by leveraging visual representations and semantic reasoning. Their model interprets driving scenarios and predicts potential dangers by analyzing scene context. Our work builds upon these efforts by employing a multi-agent vision language system designed to reason about scene context, identify anomalies, and recognize hazards in previously unseen scenarios. Instead of relying on multimodal sensor fusion, we focus on vision language models such as ViLA, OmniVLM, and GPT-4V to interpret complex visual and textual information, enabling a more adaptable and generalized hazard pipeline.

Xiao et al. explore improvements made to hazardous driving in autonomous vehicles, mainly in the events of near-miss scenarios \cite{9921988}. With the use of a kinematic-based framework, the authors detect and calculate both evasive maneuvers along with safe distances when identifying potential hazards. The paper mainly highlights the inconsistent dataset labeling methods which they address by introducing a set of safety metrics that detect evasive maneuvers, relevant to the driver state change aspect of the COOOL benchmark challenge. Relative to the benchmark dataset, this framework has some limitations as they rely on the kinematics of a vehicle which can limit the integration of high-level contextual reasoning like understanding the vehicle's visual environment. Another limitation is the GTACrash dataset itself which lacks realism and diverse hazards scenarios in real-world conditions.

Pinggera et al. \cite{pinggeralost} introduce a stereo vision based method to detect small road obstacles, addressing an important challenge in autonomous driving safety. The study focuses on hazards as small as 5 cm from a distance of 20 meters, highlighting the challenges of identifying small but potentially dangerous objects. The researchers presented the Lost and Found dataset, a collection of over 2,000 annotated stereo frames tailored for hazard detection in diverse road scenarios. The introduction of the Lost and Found dataset plays a valuable role when used for training and evaluating models. The stereo vision based detection approach achieves a high accuracy score in detecting small hazards with low false positive rates, showing promising results in comparison to other stereo-based approaches. However, its reliance on stereo vision restricts accuracy in long range or low texture areas and the dataset lacks temporal reasoning. This is because it focuses only on static frames rather than dynamic hazards. Importantly, our approach expands beyond this method by using vision-language reasoning for more complete contextual understanding and description. 

Our research expands on the above research ideas, implementing a multi-agent system that leverages zero-shot vision-language reasoning to interpret scene context, identify potential anomalies, and localize hazards without requiring additional training on predefined categories. These approaches are specifically designed to address rare and previously unseen scenarios, ensuring that autonomous systems can effectively handle rare, new, and unconventional driving hazards.

\begin{figure}
    \centering
    \includegraphics[width=0.5\textwidth]{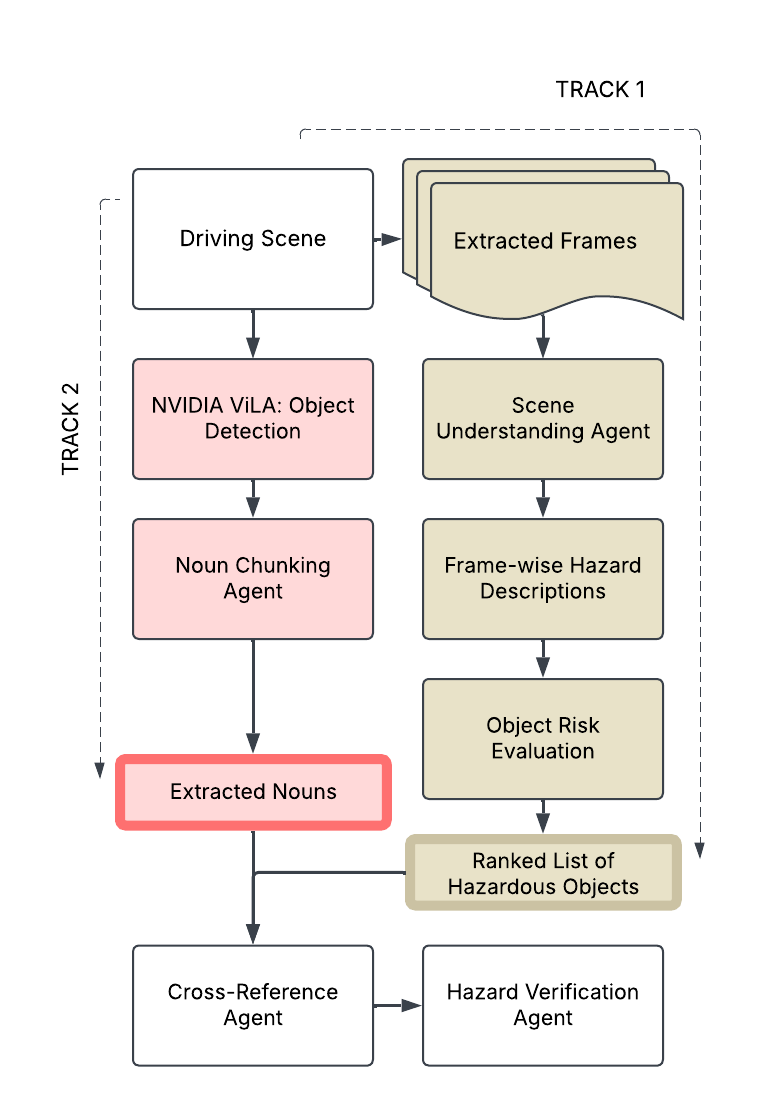}
    \caption{Our pipeline consists of two parallel tracks, both leveraging vision-language models (VLMs) and large language models (LLMs) to analyze driving scenes. One track focuses on generating detailed linguistic scene descriptions, while the other extracts key object nouns for structured representation. Through iterative refinement, these tracks produce a comprehensive list of elements most likely to be present in the scene. A zero-shot hazard verification agent then evaluates these elements to detect hazardous and anomalous objects with high precision.}
    \label{fig:earlydetection}
\end{figure}

\section{Methodologies}

Our approach to open-world driving hazard recognition follows a structured, multi-step pipeline with two parallel processes that cross reference each other for efficient hazard detection and captioning in driving scenes. An overview of this process is illustrated in Figure~\ref{fig:earlydetection}. The original COOOL Challenge consisted of three tasks: 
\begin{enumerate}
    \item \textbf{Hazard Identification:} Selecting the bounding-box-ID of hazards
    \item \textbf{Hazard Captioning:} Generating short captions of the identified Hazard
    \item \textbf{Driver Reaction:} Identifying when the driver reacts to the hazard in the video scene. 
\end{enumerate}
In this research, we will focus only on the second task: Hazard Captioning. This approach aligns with our COOOLER benchmark annotations, which are specifically designed for hazard captioning. By tailoring the annotations to these two tasks, we ensure a more standardized evaluation framework, improving model performance assessment and enabling a more reliable comparison of VLMs and LLMs in hazard detection. 


\subsection{Track 1}

The first track focuses on hazard identification and ranking based on how hazardous the object is to the scene through VLMs and LLMs. Given a driving scene video from the dataset, we extract \textit{N} frames at equal intervals \textit{S}, where \textit{N} and \textit{S} are dependent on the video length to ensure comprehensive temporal representation. For our experiments, we set \textit{N} = 25 for a 10-second video recorded at 30 fps (300 frames total), corresponding to an interval of every \textit{S} = 12 frames. This selection provides sufficient contextual diversity across the clip while maintaining computational efficiency\footnote{The implications of this decision are discussed in the Future Research section.}.  

For each extracted frame, we query a Scene Understanding Agent, implemented using a VLM, in which we utilized NexaAI's OmniVLM and the following prompt: 

\textbf{VLM's Prompt:}
\begin{quote}
\textit{“You are an autonomous vehicle looking to detect hazards or anomalies on the road. A hazard or anomaly could be an animal, person, debris or anything that is blocking the road. List and describe all the hazards and anomalies in the image.”}
\end{quote}

This results in \textit{N} hazard object descriptions for each video sequence, each detailing potential risks within the scene (e.g., \textit{“white van driving at high speed,” “pedestrian crossing unexpectedly,” “stop sign partially obstructed”, etc.}). These descriptions are then inputted into a LLM, in our case OpenAI's GPT-4o-mini, to extract all potentially hazardous objects and to rank them based on what is deemed most hazardous and least hazardous according to the model through the following prompts:

\textbf{System Prompt:}
\begin{quote}
\textit{"You are an AI assistant that identifies hazards and anomalies within descriptions of frames from a traffic scene video"}
\end{quote}

\textbf{User Prompt:}
\begin{quote}
\textit{“List1: [List of Descriptions], \newline 
List out all the potential hazardous objects and anomalies from most hazardous to least hazardous in the traffic scene and give me very short description of what the object is doing.”\footnote{In this and following prompts, we omit minor prompt details about the format / structure of the output, as this is implementation-specific.}}
\end{quote}

This then outputs a ranked list which we refer to as \textit{Ranked Hazards Set (RHS)}. This represents the prioritized hazardous elements in the scene.

\subsection{Track 2}

he second processing pathway conducts a video-wide object detection to comprehensively compile scene elements. Each video is processed using a video understanding agent (ViLA with a temperature of 0.2), which repeatedly queries the video 20 times using the prompt below in order to fully extract all details and objects of the video scene. The low temperature ensures consistent and deterministic responses, preventing unnecessary variation across iterations while enabling thorough extraction of all scene details and objects:

\textbf{VLM's Prompt:}
\begin{quote}
\textit{“What objects are on the road (provide your answer as a list separated by commas)”}
\end{quote}

This results in 20 sentence outputs, each of which are converted into a python list using the noun-chunking process provided by spaCy \cite{spacy2}, a free open source python library used for advanced Natural Language proccessing such as extracting specific nouns within sentences. These lists are then inputted into ChatGPT 4o-mini along with the following system and user prompts:

\textbf{System Prompt:}
\begin{quote}
\textit{``You are an expert evaluator tasked with selecting the most relevant and comprehensive list from the given options. The best list should contain the most important elements while avoiding redundancy and irrelevant details."}
\end{quote}

\textbf{User Prompt:}
\begin{quote}
\textit{``Here are multiple lists of objects detected in an egocentric view: 
 [Total list of objects]"\newline
``Please choose the best list based on relevance, completeness, and clarity. The ideal list should include the most important and contextually appropriate elements while avoiding duplicates and unnecessary items. Return only the best list. Also remove any articles (a, the, an) from the list."}
\end{quote}

The output from the LLM is expected to produce an output listing out all the detected objects within the video scene (e.g., \{“car”, “dog”, “stop sign”, “fire hydrant”, “cyclist”\}) We will refer to this output as \textit{All Elements Set (AES)}.


\begin{figure*}
    \centering
    \includegraphics[trim={.75cm .5cm .75cm 0cm}, clip, width=0.95\textwidth]{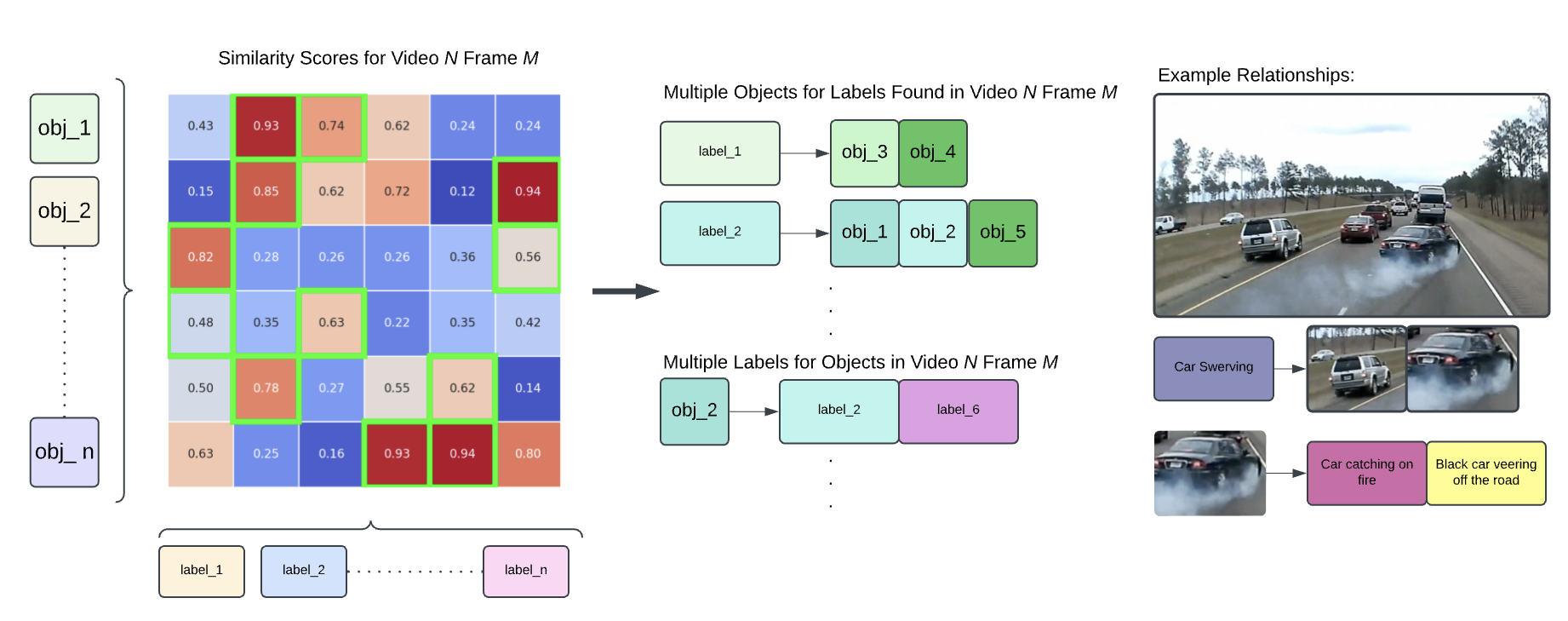}
    \caption{This figure represents a visualization of our object detection process, where object snippets are compared to label embeddings using OpenAI's CLIP model. The heatmap displays similarity scores, with each row representing an object snippet and each column corresponding to a potential label. The top 10th percentile of similarity scores (highlighted in green) is selected to identify the most relevant label associations. In the final stage, flagged elements that share common labels are grouped together, enabling more accurate object detection and classification.}
    \label{fig:coool_overview}
\end{figure*}

\subsection{Final Hazard Identification and Verification}

The final stage of our pipeline intersects these two parallel pathways that output \textit{RHS} and \textit{AES} to refine hazard identification. At this stage, we ask an LLM such as GPT-4o-mini to cross-reference the two lists, prompting it by the following:

\textbf{System Prompt:}
\begin{quote}
\textit{``You are an AI assistant that cross references two lists."}
\end{quote}

\textbf{User Prompt:}
\begin{quote}
\textit{``List1: [RHS], List2: [AES] \newline
Cross reference list1 and list2 and output a third list that ranks the common objects as shown in list2. Each index in the list should be a very short sentence description of the object."}
\end{quote}

This step filters out hazards that were inferred but not explicitly detected in the scene. The resulting ranked list is referred to as \textit{Critical Object Set (COS)} which contains objects that are both hazardous and present in the driving scene. 
After \textit{COS} is created, we further query GPT-4o-mini with the following prompt: 

\textbf{System Prompt:}
\begin{quote}
\textit{``You are an AI assistant that identifies the anomaly such as animals, birds, pedestrians and debris on the road within a list of objects belonging to a traffic scene."}
\end{quote}

\textbf{User Prompt:}
\begin{quote}
\textit{``List of Traffic Objects: [COS] \newline
Of the listed traffic objects and their descriptions, what are the anomalies / anomaly?"}
\end{quote}
 
The resulting is a set that lists anywhere from 1 or multiple identified anomalous hazards and their descriptions in case of a traffic scene containing multiple objects that pose a threat to the traffic scene. We will call this set as the \textit{Anomalous Object Set (AOS)}

To finalize hazard verification, we utilized OpenAI’s CLIP model alongside the available bounding box data. For each frame in every video, we extract \textit{snippets}—image regions corresponding to specific areas on the screen, as defined by the bounding box data. However, to ensure meaningful comparisons, \textit{snippets with insufficient resolution }
(i.e., width $< 175$ pixels, height $< 175$ pixels, or area $\leq 35,000$ pixels$^2$) are ignored to prevent unreliable matches due to low-quality visual data.

Each valid snippet is then compared against the descriptive labels in the \textit{COS}. Running these comparisons through CLIP, we compute a similarity score for every snippet-label pair, producing a 2D matrix where each row represents a challenge object (extracted snippet) and each column corresponds to a \textit{COS} label.

The similarity scores range between $0$ and $1$, indicating the degree of correlation between the extracted visual snippet and its associated textual description. To determine object detection, we select the top 10th percentile of similarity scores in each row: if an element falls within this top percentile, it is marked as detected.

Among these detected objects, we cross-reference with the Anomalous Hazard Set to identify specific snippets that correlate with marked hazards. This process allows us to track which detected objects are recognized as hazardous and the specific frames in which they appear, ensuring precise localization of anomalous hazards within the scene.

\section{Experiments and Results}

In our implementation of the above pipeline, we used OmniVLM in Track 1, ViLA in Track2, and GPT-4o-mini as our LLM. We first preprocessed and annotated the COOOL dataset  to form COOOLER, which we used to assess the pipeline's performance.

\subsection{COOOLER (Challenge Out-Of-Label with Expanded Representations}


The Challenge of Out-of-Label with Expanded Representations (COOOLER) dataset extension builds upon the COOOL benchmark, aiming to enhance the quality and effectiveness of hazard detection in autonomous driving systems and providing new methods of model evaluation. COOOLER utilizes the 200 short video clips of COOOL \cite{alshami2024coool}, which features various traffic hazards and anomalies, but updates the videos through removal of watermarks and video denoising and deblurring using NAFNet \cite{Hatami_2025_WACV}. It also includes a method for open-set evaluation, which includes hazard descriptions via freeform human annotations for the hazardous objects which are then assessed through cosine similarity and embedding-based caption analysis. In addition to open-set annotation, our enhancements to input videos included:
\begin{enumerate}
    \item \textbf{Watermark Removal} – We manually removed watermarks from the videos using an online watermark remover\footnote{https://online-video-cutter.com/remove-logo}, ensuring that artificial distractions do not interfere with hazard detection.
    \item \textbf{Denoising through NAFNet} – We first deblurred and denoised each frame of a video using NAFNet \cite{chen2022simple} to remove sensor noise and compression artifacts, enhancing the clarity of key traffic scene features. This improves Vision-Language Models (VLMs) by ensuring cleaner inputs, leading to more accurate feature extraction and scene understanding.
    \item \textbf{Deblurring through NAFNet} – Next, we applied deblurring to restore sharpness and reduce motion blur caused by vehicle movement or camera shake. This enhances fine details, allowing VLMs to extract more precise visual features, improving object recognition and spatial understanding in traffic environments.
\end{enumerate}

By following this procedure, we expect to enhance visual quality, reducing artifacts that could mislead Vision-Language Models.

\subsection{Evaluation Methods}

This benchmark aims to address hazard captioning tasks through the use of cosine similarities and caption-based embeddings. We had human annotators provide two key annotations for each of the 200 video scenes:

\begin{enumerate}
    \item \textbf{Hazardous Object Label} – A concise tag identifying the hazard, such as \textit{“Cat”}
    \item \textbf{Hazard Description} – A short caption describing the event, like \textit{“A cat crossing the road”}
\end{enumerate}

These annotations serve as reference points for evaluating model performance, ensuring a clear standard for what constitutes a correct detection. To maximize generalization, we designed captions and labels to be as flexible and inclusive as possible (open-set), accounting for variations in phrasing.

Our pipeline produces a Hazard Description \( A\), and we take its cosine similarity with the human-annotated description of the  Hazardous Object \( B \) to produce a result between 0 and 1 for each video \( i \):

\begin{equation}
S_i = \cos(\theta) = \frac{A \cdot B}{\|A\| \|B\|}
\end{equation}


We used cosine similarity to evaluate the semantic closeness between predicted and ground truth captions, ensuring models are judged on understanding rather than exact wording. Unlike string-matching metrics, cosine similarity recognizes semantically equivalent descriptions, reducing penalization for minor phrasing differences. This improves benchmarking by providing a more accurate assessment of a model’s ability to describe hazards, ensuring evaluation reflects true comprehension rather than adherence to predefined labels.

Our benchmark defines a successful detection as a cosine similarity score above 0.80 for Hazard Descriptions. This threshold ensures that predicted hazard labels and descriptions closely match human annotations in meaning, even if phrasing differs; we find that this threshold indicates strong semantic alignment, meaning the model accurately captures the essential information about the hazard.


\subsection{BESM: Balanced Extremes Similarity Metrics}

The Balanced Extremes Similarity Metric (BESM) will take the average of the maximum and minimum similarity scores within the Hazard Descriptions for each video \( i \) in order to penalize having multiple outputs that may be incorrect within the categories. We then take the average similarity scores for each video over the entire dataset in order to get the BESM of the pipelines performance with the COOOLER Benchmark:

For video \( i \):

\[
\bar{S}_i^{\text{desc}} =
\begin{cases} 
\frac{\max(S_i^{\text{desc}}) + \min(S_i^{\text{desc}})}{2} & \text{if } |S_i^{\text{desc}}| > 1 \\[10pt]
S_i^{\text{desc}} & \text{if only one unique value}
\end{cases}
\]

\[
BESM = \frac{1}{N} \sum_{i=1}^{N} \bar{S}_i^{\text{desc}}
\]


\subsection{SAM: Similarity Average Metric}
The Similarity Average Metric (SAM) is a straightforward evaluation method that computes an average similarity score across multiple components, ensuring that all similarity values contribute proportionally to the final metric. SAM aggregates description similarity scores, normalizing them based on their respective counts. This approach provides a comprehensive measure that captures variations in similarity across different aspects of the evaluation, avoiding artificial score inflation or arbitrary thresholding. SAM is expressed as:

\begin{equation}
    \bar{S}_i^{\text{desc}} = \frac{\sum_{j=1}^{m} S^{desc}_{ij}}{m }
\end{equation}

\begin{equation}
    SAM = \frac{1}{N} \sum_{i=1}^{N} \bar{S}_i^{\text{desc}}
\end{equation}

\smallskip 

\noindent where \( S^{desc}_{ij} \) are the description similarity scores (total \( m \) scores).

A key strength of SAM is that it preserves granularity in similarity scores, allowing for a more nuanced representation of similarity relationships. Unlike methods that impose hard cutoffs—potentially exaggerating high scores while disregarding small but meaningful variations—SAM ensures that both low and moderate similarity values contribute meaningfully to the overall evaluation. However, in some cases, SAM may overemphasize superficial word similarities, even when these matches are not semantically significant. This can lead to an unintended bias toward surface-level matches, potentially reducing its effectiveness in prioritizing deeper, more contextually relevant relationships. Given these characteristics, SAM provides a balanced yet sensitive metric that highlights all levels of similarity while maintaining fairness in representation.

Results for both metrics on the COOOLER dataset are provided in Table \ref{table1}.

\section{Discussion and Analysis}

\begin{table}
    \centering
    \renewcommand{\arraystretch}{1.2} 
    \setlength{\tabcolsep}{12pt} 
    \caption{Final Results of Our Pipeline on COOOLER}
    \begin{tabular}{@{}lc@{}} 
        \toprule
        \textbf{Category} & \textbf{Value} \\ 
        \midrule
        BESM   &  0.3922\\ 
        SAM    & 0.3922 \\ 
        
        \bottomrule
    \end{tabular}
    \label{table1}
\end{table}


\subsection{Strength of the Pipeline}
This multi-stage validation process, combining linguistic reasoning through VLMs and visual verification through CLIP, produces accurate hazard detection with semantic descriptions. We were able to achieve a score of 0.3922 on both BESM and SAM. The models exhibited strong contextual understanding and reasoning, recognizing hazards beyond simple object detection. Additionally, cross-referencing and ranking mechanisms refined outputs by aligning results from both tracks, filtering inconsistencies, and prioritizing critical hazards, making the system highly applicable to real-world autonomous driving.

Though evaluating on a different set of metrics, at the time of writing, the initial COOOL challenge leaderboard has an average score among top 15 of 0.386 \cite{kaggle_coolwacv25} with the best score being 0.573 \cite{Duong_2025_WACV}. We do not intend to make any direct comparison here, but simply to state that anomaly detection on this dataset is a challenging task; the best models (even on different metrics) are not achieving ``near-perfect" performance, leaving significant room for future research. 

\subsection{Limitations within the Pipeline}

The pipeline fails to detect small hazards within some of the videos, but when it does detect hazards, it does so with good semantic performance. A key observation from our experiments is that there were multiple instances where the pipeline generated no output which led to low scores for BESM and SAM. This primarily occurred due to two reasons: 

\begin{itemize}
    \item \textbf{Failure to Detect Hazardous Objects: } There is sometimes failure in the merging of Track 1 and Track 2 outputs, where the Vision-Language Model (VLM) or Large Language Model (LLM) did not detect a hazardous object. This fault could have happened for two reasons: 

    \begin{itemize}
        \item The VLMs were not able to detect the hazardous object within the video sequence.
        \item The LLM was not able to reason for any of the listed hazardous objects to be hazardous enough. 
    \end{itemize}

    \item \textbf{Object Size Limitations: } Some hazards were not detected because the bounding box annotations of the hazardous objects were too small, leading to their exclusion from our pipeline. These limitations highlight the need for refining bounding box processing within our pipeline in order to pass through smaller objects.
\end{itemize}

\subsection{Evaluation of COOOLER}
The COOOLER Benchmark provided a structured evaluation of our pipeline's hazard detection capabilities, offering insights through multiple metrics. BESM penalized inconsistent outputs, revealing stability issues but being sensitive to outliers. SAM provided a balanced average similarity but struggled to highlight peak performance. However, the evaluations also exposed limitations, such as failures in detecting small or occluded hazards and score fluctuations due to minor linguistic variations.

Additionally, the driving scenes in our dataset occasionally featured scenarios where no true hazard was present or where the hazard was insignificant to the driver. For example, in  \textit{video\_0005}, a scene captures a driver on the freeway passing an exit ramp, where a small piece of debris is present on the ground of the ramp. While technically an object in the environment, such instances do not pose a direct threat to the driver. Our pipeline is designed to intentionally ignore hazard objects that are too small to be meaningfully detected, prioritizing actionable hazards over minor environmental anomalies. However, this results in penalization under the COOOLER evaluation metric, as the final scores are averaged across all videos, including those where small, non-threatening objects are present but not flagged by the system. This highlights a tradeoff between filtering out inconsequential hazards and maintaining high recall in benchmark evaluations. Despite these challenges, COOOLER effectively identified strengths and weaknesses in our pipeline, emphasizing the need for improvements in bounding box processing, scene context modeling, and evaluation flexibility to enhance detection accuracy and robustness.

\section{Future Research}

\subsection{Pipeline Refinement}
While our pipeline significantly improves hazard detection accuracy, several challenges remain. Failures in merging detection tracks and difficulties recognizing small or occluded objects impacted performance. Future research should refine bounding box processing, cross-referencing strategies, and motion-based reasoning to enhance object recognition and reduce false negatives. Additionally, further prompt engineering to increase the contextual understandings of VLMs and LLMs could also be quite beneficial. 

\subsection{Evaluating Additional Vision Language Models and Large Language Models}

While our pipeline successfully leveraged OmniVLM and ChatGPT-4o-mini as our VLM and our LLM, further exploration of alternate models could help with the evaluation of detection accuracy and reasoning capabilities between these models. Different combinations of models can be used to evaluate their reasoning and to also help address the limitations where the pipeline fails to reason for and detect hazardous objects in the COOOLER dataset's video sequences. Exploration of these models could also be beneficial in evaluating their applicability within autonomous driving scenes. 


\subsection{Further Exploring the Other Two Tasks of COOOL}
Our research has focused on hazard captioning, but the original COOOL Benchmark also includes the identification of the bounding box labels of the hazardous object and Driver Reactions task determining when a driver responds to a hazard. Future work should explore temporal modeling approaches such as transformers or recurrent neural networks to track hazard evolution and predict driver reactions. Integrating sequential reasoning would enhance our ability to model cause-and-effect relationships within dynamic driving scenarios, further improving safety assessment in autonomous systems.

\section{Concluding Remarks}
This research advances hazard detection in autonomous driving by integrating the zero-shot reasoning of Vision-Language Models (VLMs) and Large Language Models (LLMs) for the open-set recognition of novel hazards beyond predefined categories. Through the COOOLER Benchmark, we established a structured evaluation framework, enhancing dataset quality with video preprocessing, refined annotations, and new methods for evaluation on Hazard Descriptions provided by the pipeline's output. Our multi-agent pipeline,  demonstrated strong context-aware hazard identification, improving adaptability and detection reliability, but limitations still remained particularly in handling small or occluded hazards and refining detection consistency across frames. This work lays the foundation for real-time hazard assessment, predictive autonomous systems, and safer AI-driven navigation. As vision-language models continue to evolve, their integration into autonomous driving has the potential to significantly enhance traffic safety, reduce accident risks, and improve situational awareness, bringing autonomous vehicles closer to human-level hazard perception.

\bibliographystyle{IEEEtran}
\bibliography{IEEEReferences}

\end{document}